\documentclass[runningheads]{llncs}

\usepackage[T1]{fontenc}

\usepackage{graphicx}

\usepackage{hyperref}
\usepackage{color}
\usepackage{amsmath}
\usepackage{booktabs}
\usepackage{makecell}
\usepackage{multirow}
\usepackage{fontawesome5}
\usepackage{float}

\begin{document}

\author{Roseval Malaquias Junior\inst{1} \and
Giovana Kerche Bonás\inst{1} \and
Thales Sales Almeida\inst{1} \and
Hugo Abonizio\inst{1} \and
Thiago Laitz\inst{1} \and
Ramon Pires\inst{1} \and
Marcos Piau\inst{2} \and
Celio Larcher\inst{2} \and
Rodrigo Nogueira\inst{1}
}

\institute{Maritaca AI \and Jusbrasil}

\authorrunning{R.\ Malaquias Jr.\ et al.}
\titlerunning{Prosa}

\title{Prosa: Rubric-Based Evaluation of LLMs on Real User Chats in Brazilian Portuguese}

\maketitle
\begin{abstract}
Rankings produced by holistic LLM-as-a-judge scoring are sensitive to the bias of the chosen judge model. We show that switching to binary rubric scoring with multi-judge filtering removes this sensitivity: decomposing the judgement matters more than the judge model itself. To support this claim, we introduce Prosa\footnote{All code, benchmark, and rubrics: \url{https://github.com/maritaca-ai/Prosa}.}, the first real user multi-turn Brazilian Portuguese chat benchmark: 1{,}000 WildChat conversations scored by three judges from three model families on 16 models. Under filtered rubric scoring the three judges agree on every one of the 16 ranks, whereas under holistic scoring they agree on only 7 of 16. Additionally, the rubric filtering pipeline increases the average score gap between neighbouring models by 47\%, thereby improving Prosa's discriminative power. Evaluating a new model on Prosa costs approximately \$2.1 when using Gemini 3 Flash as the judge. We release the benchmark and the filtering code to ensure that future models can be assessed under identical conditions. These artifacts also make our rubric-based scoring method reusable beyond Prosa, supporting other open-ended evaluation settings.
\end{abstract}

\section{Introduction}

Automatic evaluation of open-ended language generation has long relied on n-gram overlap metrics such as BLEU and ROUGE. Because they reward only lexical overlap with a reference, these metrics correlate weakly with human judgement on open-ended tasks~\cite{liu2016how,ma2019wmt19,reiter2018bleu}. The failure is sharpest near the top of the quality range: they separate bad from good systems but fail to rank good systems among themselves~\cite{sellam2020bleurt}. LLM-as-a-judge evaluation addresses this limitation, reaching inter-annotator agreement on par with expert humans~\cite{NEURIPS2023_91f18a12}. Modern benchmarks such as MT-Bench~\cite{NEURIPS2023_91f18a12}, AlpacaEval 2 LC~\cite{dubois2024lengthcontrolled}, Arena-Hard-Auto~\cite{li2025arenahard}, and WildBench~\cite{lin2025wildbench} converged on this paradigm, scoring responses holistically as a single $[1, 10]$ score or a pairwise preference. In both cases, the judge must collapse multiple dimensions of quality into a single decision, leaving the score dependent on the judge's bias. Holistic scoring is convenient but opaque at the criterion level, and we show that this dependence makes the resulting ranking sensitive to the choice of judge.

This progress has been heavily centred on English. Existing Brazilian Portuguese benchmarks are dominated by academic multiple-choice exams~\cite{silveira2018advances,nunes2023evaluating,almeida2023bluex} and translated variants of English tasks~\cite{correa2026tucano2,almeida2025poetav2}; none of them observe how Brazilian users actually interact with LLMs for their own purposes. A benchmark derived from naturally occurring interactions surfaces the topics, registers, and patterns of use that matter in Brazil, and evaluates model behaviour under conditions closer to deployment than exam-based or translated benchmarks can offer.

We introduce Prosa\footnote{\textit{Prosa} is the Brazilian Portuguese word for informal conversation.}, the first real user multi-turn Brazilian Portuguese chat benchmark with rubric-based LLM-as-judge evaluation. It rests on three design choices. First, it derives its 1{,}000 conversations from WildChat~\cite{zhao2024wildchat} with the original multi-turn context preserved. Second, each response is scored against a set of binary pass/fail rubrics adapted from RRD~\cite{shen2026rethinking} and aligned with the decomposed-requirement protocol of InfoBench~\cite{qin2024infobench}. Third, a multi-judge post-hoc filtering pipeline removes low-quality rubrics before scoring.

The central empirical finding is that decomposing the judgement (into atomic binary verdicts rather than a single score) matters more than the judge model itself: three judges from three model families agree on every one of the 16 ranks under filtered rubric scoring, while the same judges disagree on multiple rank positions under holistic scoring. The filtering pipeline alone raises the average gap between neighbouring models by 47\%, sharpening Prosa's discriminative power.

Two practical implications follow. First, because each rubric reduces the judge's task to a single binary verdict, the methodology does not demand a frontier judge: Gemini 3 Flash at \$2.1 per candidate reproduces the ranking obtained with the most expensive judge in our suite, lowering the cost barrier for large-scale evaluation. Second, the per-rubric verdicts provide a more discriminative, fine-grained quality signal than a single holistic score, which may make the same rubric set usable for reward modelling in addition to evaluation.

Section~\ref{sec:related} positions Prosa in the three lines of literature that motivate its design. Sections~\ref{sec:construction} and~\ref{sec:protocol} describe the benchmark construction and the evaluation protocol. Section~\ref{sec:results} reports the empirical validation and the final model ranking. Section~\ref{sec:conclusion} concludes.

\section{Related Work}\label{sec:related}

\subsection{Real user open-ended benchmarks}\label{sec:rw-realuser}

Open-ended LLM evaluation with LLM-as-judge scoring was largely established by MT-Bench~\cite{NEURIPS2023_91f18a12} (80 expert-written multi-turn prompts scored on $[1, 10]$) and AlpacaEval 2 LC~\cite{dubois2024lengthcontrolled} (pairwise single-turn evaluation over the AlpacaFarm aggregation, with regression-based length-bias control).

A second line builds benchmarks directly from user interactions. Arena-Hard-Auto~\cite{li2025arenahard} curates 500 challenging single-turn prompts from Chatbot Arena logs and scores responses pairwise against a strong baseline. WildBench~\cite{lin2025wildbench} samples multi-turn tasks from WildChat and scores responses either pairwise (WB-Reward) or as a single $[1, 10]$ score in which a task-specific checklist is provided to the judge as contextual guidance (WB-Score). MultiChallenge~\cite{sirdeshmukh2025multichallenge} targets realistic multi-turn conversations but synthesises them with persona seeds and human edits rather than drawing them from real user logs, scoring responses against instance-level binary rubrics.

Prosa adapts this real user line to Brazilian Portuguese with three scoring-side differentials. First, it scores responses with binary pass/fail rubrics rather than holistic scoring (motivated in Section~\ref{sec:rw-rubrics}). Second, it introduces an automatic multi-judge post-hoc filtering pipeline (Section~\ref{sec:filter}) that removes low-quality rubrics before scoring. Third, because each rubric reduces the judge's task to a single binary decision, weaker and cheaper judges suffice: evaluating a candidate with Gemini 3 Flash costs approximately \$2.1, an order of magnitude below the \$20 per candidate that Arena-Hard-Auto~\cite{li2025arenahard} reports. Table~\ref{tab:text_benchmarks} summarises how Prosa relates to existing benchmarks.

\begin{table}[htb]
\centering
\caption{Open-ended LLM-as-judge evaluation benchmarks. Prosa is the only one combining real user data, binary rubrics, and automatic rubric filtering.}
\label{tab:text_benchmarks}
\renewcommand{\arraystretch}{1.3}
\setlength{\tabcolsep}{5pt}
\resizebox{\linewidth}{!}{%
\begin{tabular}{l c c c c c c}
\toprule
\multirow{2}{*}{Benchmark} & \multicolumn{3}{c}{\textbf{Prompt data}} & \multicolumn{3}{c}{\textbf{Scoring protocol}} \\
\cmidrule(lr){2-4} \cmidrule(lr){5-7}
 & Source of prompts & Real user & Multi-turn & Holistic scoring & Binary rubrics & Automatic rubric filtering \\
\midrule
MT-Bench       & Expert-written           & \faTimes   & \faCheck & \faCheck & \faTimes & \faTimes \\
AlpacaEval 2 LC & AlpacaFarm eval set     & \faTimes   & \faTimes & \faCheck & \faTimes & \faTimes \\
Arena-Hard-Auto & Chatbot Arena, WildChat & \faCheck   & \faTimes & \faCheck & \faTimes & \faTimes \\
WildBench       & WildChat                & \faCheck   & \faCheck & \faCheck & \faTimes & \faTimes \\
MultiChallenge  & Synthetic + human-edited & \faTimes  & \faCheck & \faTimes & \faCheck & \faTimes \\
\midrule
\textbf{Prosa (ours)} & \textbf{WildChat}  & \textbf{\faCheck} & \textbf{\faCheck} & \textbf{\faTimes} & \textbf{\faCheck} & \textbf{\faCheck} \\
\bottomrule
\end{tabular}%
}
\end{table}

\subsection{Decomposed rubric-based evaluation}\label{sec:rw-rubrics}

Holistic LLM-as-judge scoring forces the judge to integrate many evaluation dimensions into a single number, which limits reproducibility, obscures the reasoning behind the final score, and demands a strong judge to perform the integration reliably. A parallel literature addresses this by decomposing the judgement into finer-grained criteria, where each verdict is local to one criterion and therefore simpler to elicit. InfoBench~\cite{qin2024infobench} introduces the Decomposed Requirements Following Ratio, breaking each instruction into binary pass/fail requirements and computing the score as the compliance rate. FLASK~\cite{ye2024flask} and BiGGen Bench~\cite{kim2024biggen} extend this idea with skill-set-based and instance-specific rubrics on graded scales.

RRD (Recursive Rubric Decomposition)~\cite{shen2026rethinking} is the framework our scoring protocol most directly builds on. It formalises binary rubrics as functions $g_k: P \times R \to \{0, 1\}$ that map a prompt–response pair to a pass/fail verdict, and proposes a decompose-filter loop that refines an initial rubric set by splitting coarse criteria into finer ones, removing overlapping or conflicting rubrics, and flagging rubrics whose verdicts are inverted with respect to a weak reference model. The authors show that rubrics refined by this pipeline improve both judge accuracy and reward-modelling stability across several open-ended tasks, evidence that decomposed binary verdicts can serve both evaluation and reward-signal roles.

Prosa inherits two components from this literature and adds one. From InfoBench: our scoring formula (Section~\ref{sec:scoring}) is DRFR scaled to a 0--100 range. From RRD: the F.1 rubric-generation and F.3 binary-judging prompts, adapted to Brazilian Portuguese; we do not run RRD's decompose-filter loop or weight-assignment stage. Our addition is the multi-judge post-hoc filtering pipeline (Section~\ref{sec:filter}), which operates on the empirical ranking of 16 candidate models rather than on a weak reference model and, together with the real user Brazilian Portuguese application, defines what is original about Prosa.

\subsection{Brazilian Portuguese LLM evaluation}\label{sec:rw-ptbr}

Most Brazilian Portuguese LLM benchmarks focus on academic multiple-choice exams such as ENEM~\cite{silveira2018advances,nunes2023evaluating} and BLUEX~\cite{almeida2023bluex}. Open-ended LLM-as-judge evaluation in Brazilian Portuguese has emerged more recently. Sabi{\'a}-2~\cite{almeida2024sabia2} introduces BRACEval, an in-house multi-turn benchmark of 150 questions inspired by MT-Bench. OAB-Bench~\cite{pires2025oabbench} evaluates legal-essay writing on questions from the Brazilian Bar Exam, graded by an o1 judge against official reference materials.

To the best of our knowledge, no prior Brazilian Portuguese benchmark combines real user conversations, multi-turn context, and binary rubric-based LLM-as-judge evaluation. Prosa occupies this intersection by deriving its 1{,}000 conversations from WildChat, preserving their natural multi-turn structure, and scoring responses with the rubric-based protocol introduced in Section~\ref{sec:rw-rubrics}.

\section{Prosa Construction}\label{sec:construction}
\begin{figure}[H]
    \centering
    \includegraphics[width=\linewidth]{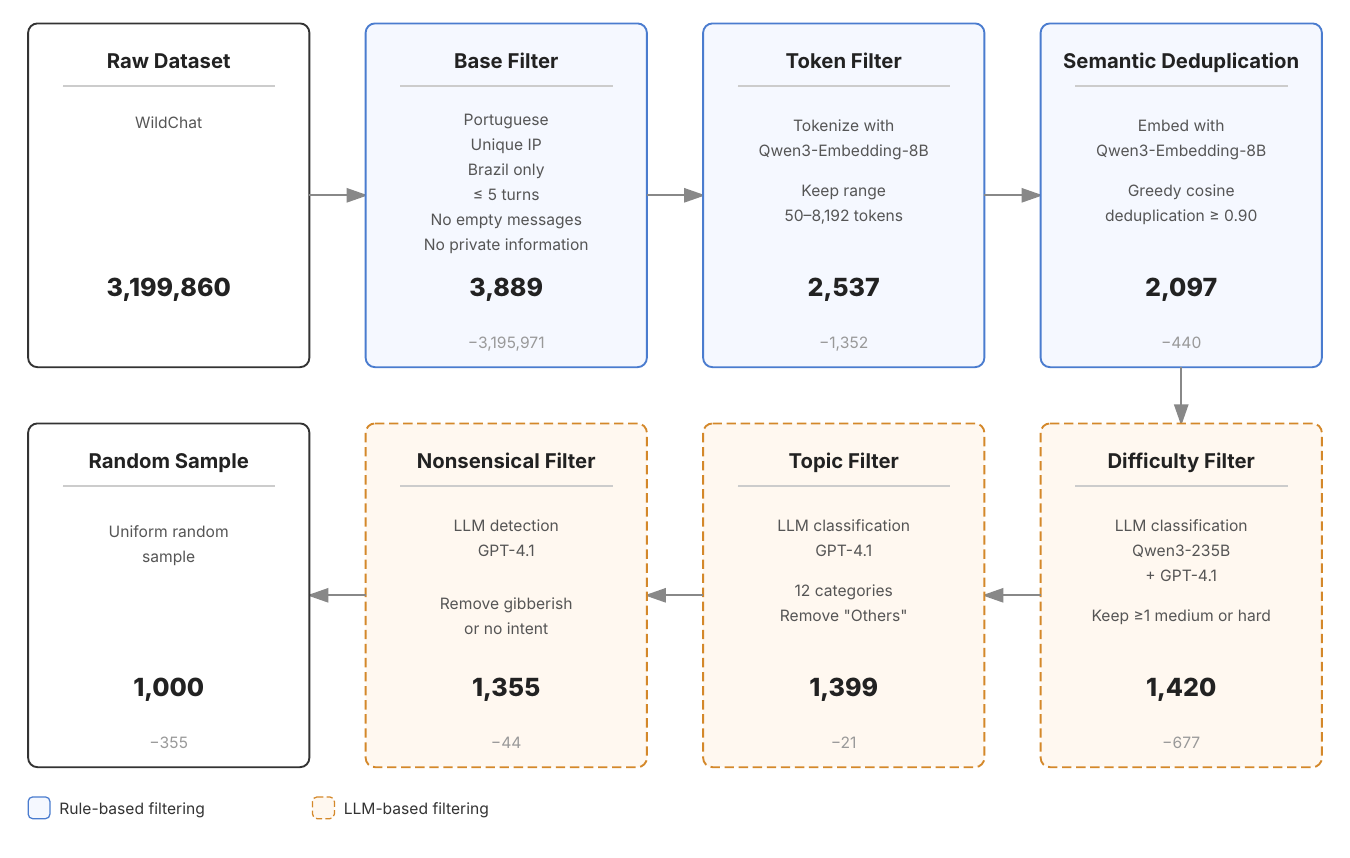}
    \caption{Conversation filtering pipeline reducing 3.2M WildChat conversations to the final 1{,}000 of Prosa. Solid borders mark rule-based filters (metadata, token length, and embedding similarity); dashed borders mark LLM-based filters (difficulty, topic, and intent clarity). A random sample finalises the benchmark.}
    \label{fig:pipeline}
\end{figure}

Prosa is constructed from the WildChat dataset~\cite{zhao2024wildchat}, which contains 3.2 million conversation logs and includes only non-toxic user interactions, as released by the authors\footnote{\url{https://huggingface.co/datasets/allenai/WildChat-4.8M}}. Real user logs frequently contain tasks whose outcomes cannot be reliably verified, or that are too simple or repetitive to discriminate between models~\cite{li2025arenahard}; our conversation filtering pipeline (Figure~\ref{fig:pipeline}) therefore retains only verifiable, non-redundant tasks written in Portuguese and originating from Brazil, reducing the corpus to 1{,}000 conversations through six successive filters and a final random sample. The topic and difficulty labels assigned during filtering also serve as metadata for the analyses in Section~\ref{sec:results-ranking}. We release the full pipeline so that the same procedure can be applied to other real user conversation logs.

\subsection{Rule-based filtering}

Three rule-based stages on metadata, length, and embedding similarity precede any LLM judgement. The \emph{Base Filter} keeps only conversations in Portuguese originating from Brazil with non-empty user messages, retains at most one per unique IP address, limits length to five user--model turns, and removes any conversation containing the \texttt{REDACTED} marker (3.2M $\rightarrow$ 3{,}889). The \emph{Token Filter} discards conversations whose total length, measured with the Qwen3-Embedding-8B tokenizer~\cite{qwen3embedding}, falls outside $[50, 8192]$ tokens (3{,}889 $\rightarrow$ 2{,}537). \emph{Semantic Deduplication} encodes each remaining conversation with the same model and removes pairs with cosine similarity $\geq 0.90$ (2{,}537 $\rightarrow$ 2{,}097).

\subsection{LLM-based filtering}

Three LLM-driven stages refine the surviving conversations, supported by prior work on LLM-based automatic annotation~\cite{NEURIPS2023_91f18a12,xu2025magpie}. The \emph{Difficulty Filter} prompts \texttt{gpt-4.1-2025-04-14} and \texttt{Qwen3-235B-A22B-Instruct-2507} to classify each task on a five-level scale (\emph{very easy} to \emph{very hard}); we retain examples where at least one judge labels the task at \emph{medium} or above (2{,}097 $\rightarrow$ 1{,}420). The \emph{Topic Filter} uses \texttt{gpt-4.1-2025-04-14} to assign each example to one of 12 categories inspired by the Magpie taxonomy~\cite{xu2025magpie}; examples that cannot be confidently assigned (labelled \emph{Others}) are removed (1{,}420 $\rightarrow$ 1{,}399). The \emph{Nonsensical Filter} uses the same judge to remove examples that lack a clear, evaluable user request (1{,}399 $\rightarrow$ 1{,}355). A final random sample of 1{,}000 examples preserves the natural distribution of topics, token lengths, and turns.

\subsection{Benchmark statistics}

\begin{figure}[H]
    \centering
    \includegraphics[width=\linewidth]{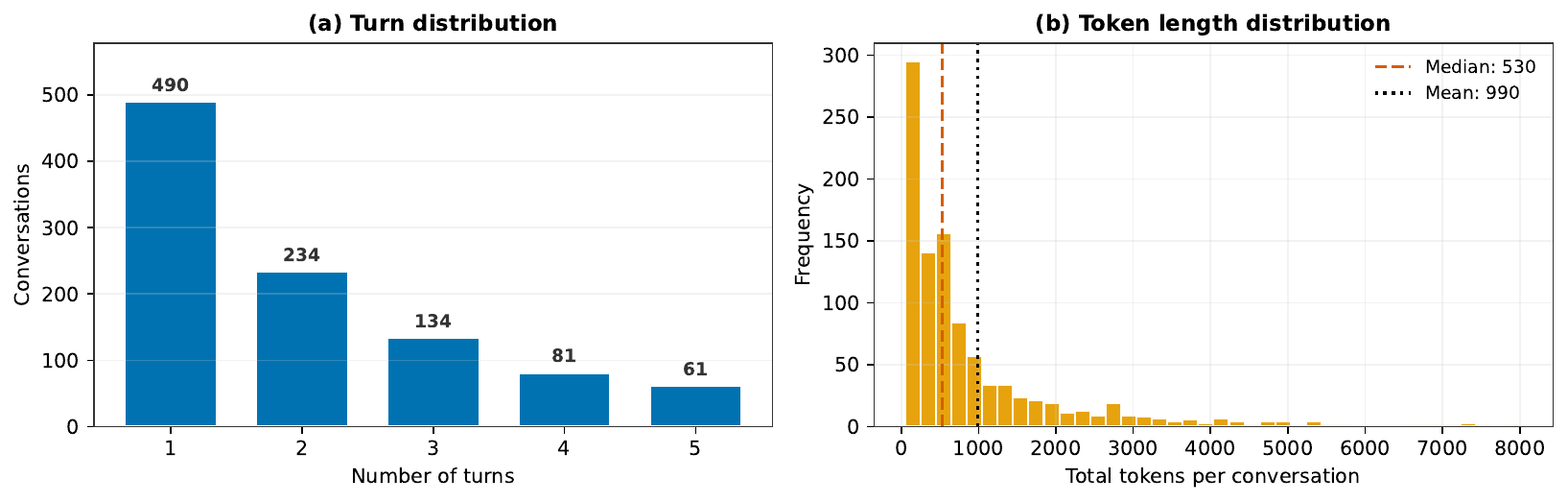}
    \caption{Conversational composition of Prosa ($n{=}1{,}000$): (a) turn count distribution ($51\%$ multi-turn, mean $1.99$ turns); (b) token counts (median $530$, mean $990$, long-tailed).}
    \label{fig:benchmark-composition}
\end{figure}

\begin{figure}[H]
    \centering
    \includegraphics[width=\linewidth]{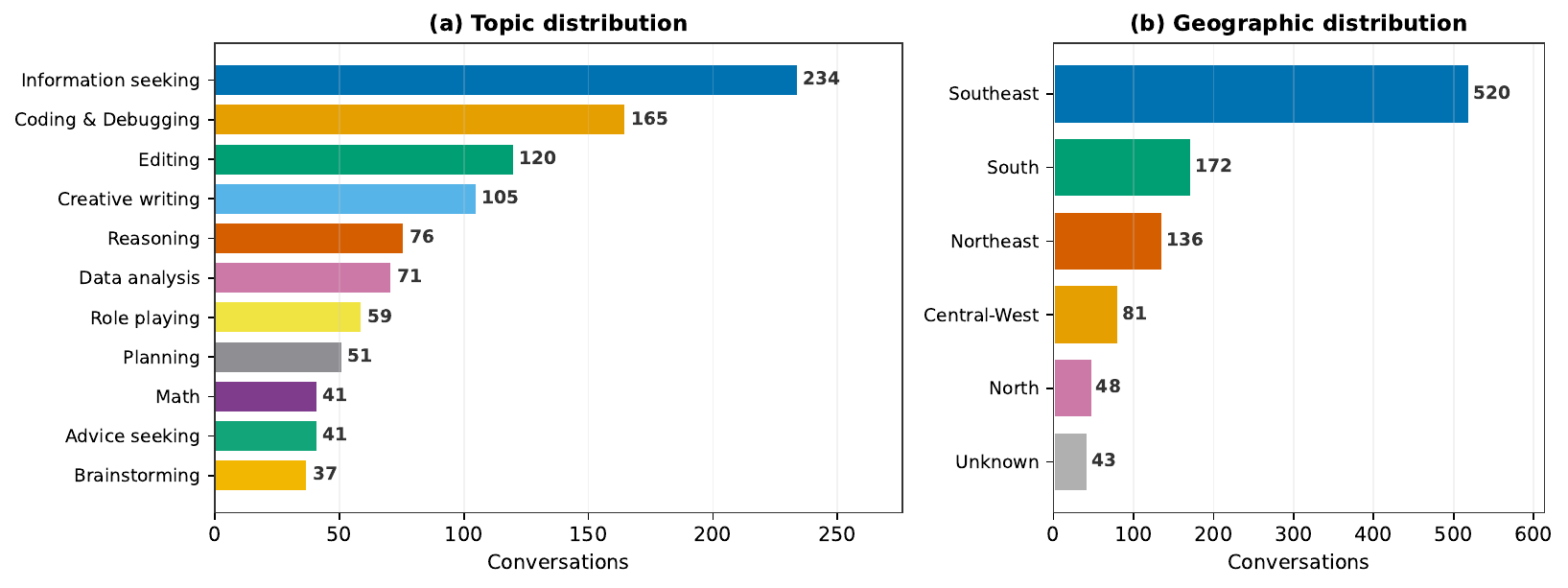}
    \caption{Topical and geographic coverage of Prosa ($n{=}1{,}000$): (a) 11 topical categories from the Magpie taxonomy~\cite{xu2025magpie}; (b) all five Brazilian regions, with a Southeast majority.}
    \label{fig:benchmark-coverage}
\end{figure}

The final 1{,}000 conversations are 51\% multi-turn (mean 1.99 user turns), with token counts ranging widely (median $530$, mean $990$); this distinguishes Prosa from single-turn benchmarks such as AlpacaEval 2 LC~\cite{dubois2024lengthcontrolled} and Arena-Hard-Auto~\cite{li2025arenahard} (Figure~\ref{fig:benchmark-composition}). Topics span 11 categories with \textit{Information seeking} the largest at $23.4\%$ and the remaining ten between $3\%$ and $17\%$; geographically, $95.7\%$ of conversations carry a state annotation covering all five Brazilian regions, with the Southeast accounting for $52.0\%$ (Figure~\ref{fig:benchmark-coverage}).

\section{Evaluation Protocol}\label{sec:protocol}

\subsection{Rubric generation}\label{sec:generation}

We adopt the F.1 rubric-generation prompt from RRD~\cite{shen2026rethinking}, applied directly to our 1{,}000 filtered conversations through a Brazilian Portuguese translation of the template. We do not run RRD's decompose-filter loop or its weight-assignment stage; our multi-judge filtering pipeline (Section~\ref{sec:filter}) plays that role.

For each conversation, the generator (GPT-5.2) is conditioned on the conversation history, the user query, and three reference responses from strong models spanning distinct families (GPT-5.2, Qwen3.5-397B-A17B, Gemini 3 Pro), and produces a set of binary pass/fail criteria covering the dimensions needed to judge a response. The resulting set comprises 12{,}920 atomic binary criteria ($g_k: P \times R \to \{0, 1\}$), with a mean of 12.9 rubrics per conversation.

\subsection{Rubric-level binary scoring}\label{sec:scoring}

The judge emits an independent YES/NO verdict for every rubric of Section~\ref{sec:generation}, and the response score is computed from these verdicts. We refer to this protocol as \emph{rubric scoring}, in contrast to the \emph{holistic scoring} used by WildBench's WB-Score~\cite{lin2025wildbench}, in which the judge reads the checklist as contextual guidance but returns a single $[1, 10]$ score.

Our judge prompt is a Brazilian Portuguese adaptation of the F.3 template from RRD~\cite{shen2026rethinking}. In a single judgement, the judge receives the conversation history (when present), the current user query, the response under evaluation, and the full rubric list, and emits one YES/NO verdict per rubric. Batching all rubrics of a question into a single judgement amortises the conversation context across rubrics, reducing inference cost.

Let $g_k(m, q) \in \{0, 1\}$ be the verdict on rubric $k$ of question $q$ for model $m$, and let $K_q$ be the number of rubrics for question $q$. The score for model $m$ pools rubric verdicts across the benchmark,
\[
    s(m) = 100 \cdot \frac{\sum_{q, k} g_k(m, q)}{\sum_q K_q}.
\]
This is InfoBench's Decomposed Requirements Following Ratio (DRFR)~\cite{qin2024infobench} scaled to a 0--100 range.

\subsection{Multi-judge rubric filtering}\label{sec:filter}

Not every rubric carries discriminative value: some are trivially satisfied or universally failed, some correlate negatively with response quality, and some flip their verdict across repeated runs at temperature $0$. We apply five post-hoc filters to remove rubrics exhibiting these pathologies, operating on the $16 \times 1{,}000$ evaluation matrix produced by the three judges of Section~\ref{sec:setup}.

Three filters use majority voting across the three judges ($\geq 2/3$ agreement) on the candidate verdicts. \emph{Trivial} marks a rubric that every one of the 16 candidates passes; \emph{impossible} marks a rubric that every candidate fails; \emph{misaligned} marks a rubric on which the two top-scoring candidates fail while the bottom-scoring candidate passes, indicating the verdict is inverted with respect to overall response quality. The misaligned filter is related to RRD's misalignment detection~\cite{shen2026rethinking}, but operates on the empirical ranking of the candidates rather than on a weak reference model.

The fourth filter, \emph{unstable}, targets non-reproducible rubrics: with Qwen3-30B as candidate and GPT-4.1 as judge, we run three independent evaluations at temperature $0$ and remove any rubric whose verdict changes across the three runs. The fifth filter operates at the question level: a question with no surviving rubrics is dropped from scoring.

Applied to Prosa, the five filters remove 4{,}512 rubrics ($34.9\%$ of the 12{,}920 originally generated) and drop 19 questions; the effective benchmark comprises 981 questions and 8{,}408 rubrics (mean 8.6 per question). Section~\ref{sec:results-validation} reports their empirical impact on inter-judge agreement, ranking stability, and discriminative power.

\subsection{Experimental setup}\label{sec:setup}

We evaluate 16 candidates from four model families (Table~\ref{tab:candidate-models}). Proprietary models are served through their providers' native APIs; open-weights Qwen models are served via OpenRouter when available, with local inference as fallback.

We use three judges spanning three model families: GPT-4.1, Gemini 3 Flash, and Sabi{\'a}-4 (Table~\ref{tab:candidate-models}). All candidate responses and judge verdicts are decoded at temperature $0$. We report three families of validation metrics in Section~\ref{sec:results-validation}. \emph{Rank stability} measures whether the three judges produce the same leaderboard: we report the Spearman correlation between each pair of judges, averaged over the three pairs, and the number of models that occupy the same rank under both judges in a pair, taking the minimum over pairs to surface the most divergent pair. \emph{Inter-judge agreement} measures how often the judges return the same YES/NO verdict for a given (model, rubric) pair: we report the fraction of such pairs on which all three judges agree. \emph{Discriminative power} measures how cleanly the protocol separates adjacent models on the leaderboard: we report the average gap between consecutive ranks and the top-to-bottom spread (rank 1 minus rank 16), each averaged over the three judges on a 0--100 scale.

\begin{table}[H]
\centering
\caption{Candidate models evaluated in Prosa, grouped by family. Responses come from the provider's native API for proprietary models, OpenRouter when available for open-weights Qwen models, and local inference otherwise.}
\label{tab:candidate-models}
\renewcommand{\arraystretch}{1.1}
\scalebox{1.0}{%
\begin{tabular}{@{}l l l@{}}
\toprule
Family & Model & Inference route \\
\midrule
OpenAI   & GPT-5.2          & OpenAI API \\
         & GPT-5 Mini       & OpenAI API \\
         & GPT-4.1          & OpenAI API \\
         & GPT-4.1 Mini     & OpenAI API \\
         & GPT-4o           & OpenAI API \\
         & GPT-4o Mini      & OpenAI API \\
\midrule
Google   & Gemini 3 Pro     & Google AI API \\
         & Gemini 3 Flash   & Google AI API \\
         & Gemini 2.5 Flash & Google AI API \\
\midrule
Alibaba  & Qwen3-235B       & OpenRouter \\
         & Qwen3-30B        & OpenRouter \\
         & Qwen2.5-7B       & OpenRouter \\
         & Qwen3-4B         & local (A100) \\
         & Qwen2.5-14B      & local (A100) \\
\midrule
Maritaca AI & Sabi{\'a}-4          & Maritaca AI API \\
         & Sabi{\'a}-3.1        & Maritaca AI API \\
\bottomrule
\end{tabular}%
}
\end{table}

Validating the benchmark with all three judges over the 16 candidates and 981 questions costs approximately \$113 (GPT-4.1), \$34 (Gemini 3 Flash), and \$66 (Sabi{\'a}-4). Evaluating a new candidate requires only one judge; the cheapest is Gemini 3 Flash at $\sim$\$2.1 per candidate. For reference, Arena-Hard-Auto~\cite{li2025arenahard} reports \$20 per candidate over 500 prompts. We release the rubric-generation, scoring, and filtering code, with the rubrics and filter outcomes frozen, so that any new candidate can be evaluated with a single judge of choice without re-running the filtering stage.

\section{Results}\label{sec:results}

\subsection{Benchmark validation}\label{sec:results-validation}

\begin{figure}[H]
    \centering
    \includegraphics[width=\linewidth]{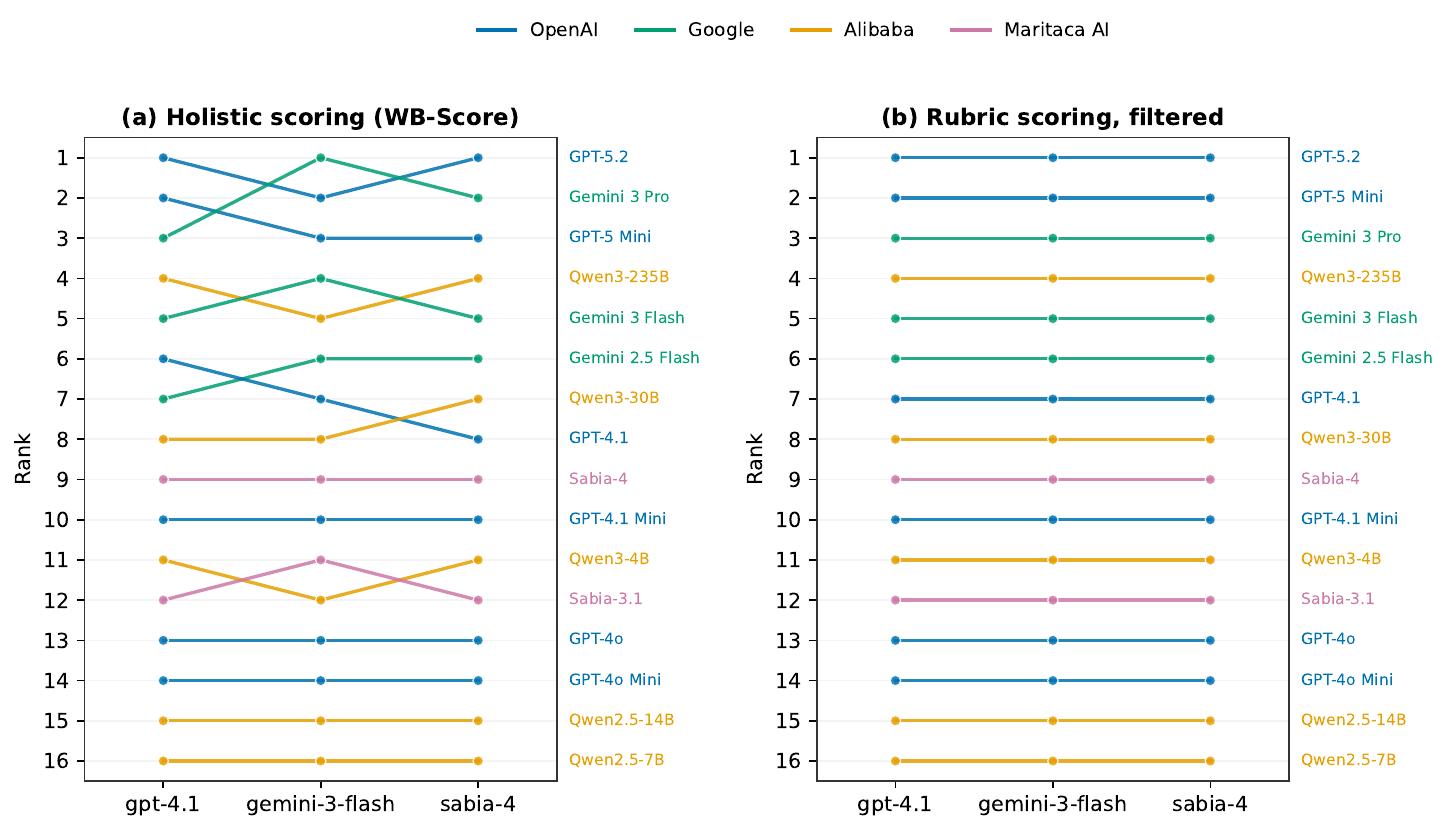}
    \caption{Candidate rank across the three judges under (a) holistic scoring and (b) filtered rubric scoring. Holistic disagreement is sharpest at the top (Gemini 3 Flash places Gemini 3 Pro first, others place GPT-5.2 first); under rubric scoring, all three judges agree on every rank.}
    \label{fig:validation-bump}
\end{figure}

We compare three protocols on the same 1{,}000 conversations and three judges: \emph{holistic scoring}, which reproduces WildBench's WB-Score~\cite{lin2025wildbench}; \emph{rubric scoring} without filtering; and \emph{rubric scoring} with the pipeline of Section~\ref{sec:filter} applied. Table~\ref{tab:validation} reports rank stability, inter-judge agreement, and discriminative power across the three protocols, using the metrics defined in Section~\ref{sec:setup}. Figure~\ref{fig:validation-bump} visualises how the rank of each candidate shifts across judges under the first and last protocols.

Under holistic scoring the three judges do not agree on the ranking: Spearman averages 0.986 but only 7 of 16 candidates occupy the same rank in the most divergent judge pair. Figure~\ref{fig:validation-bump}(a) makes the mechanism explicit: the Gemini 3 Flash judge places Gemini 3 Pro at rank 1, while GPT-4.1 and Sabi{\'a}-4 place GPT-5.2 at rank 1, with several adjacent positions also shuffled. Switching to rubric scoring largely eliminates this divergence: even without filtering, Spearman is 0.998 and 14 of 16 candidates occupy identical ranks across judge pairs; with filtering, Spearman reaches exactly 1.000 and every candidate occupies the same rank (Figure~\ref{fig:validation-bump}(b)). The contrast is already largely present without filtering, indicating that atomic binary verdicts carry most of the mechanism, while the filtering pipeline closes the remaining disagreement on two of the sixteen ranks.

The pipeline also sharpens the separation between adjacent ranks. Without filtering, the average gap between neighbouring candidates is 2.22 points and the top-to-bottom spread is 33.3 points, close to what holistic scoring achieves (2.24 and 33.5). Applying the pipeline raises the average gap to 3.26 (a 47\% increase) and expands the spread to 48.9 points. This gain in separability comes at the cost of a small drop in inter-judge agreement: unanimous YES/NO verdicts fall from 87.6\% to 83.8\% of the (model, rubric) pairs (over 12{,}920 and 8{,}408 rubrics, respectively), which is expected since the trivial filter removes precisely the rubrics on which agreement is easiest.

\begin{table}[!htbp]
\centering
\caption{Validation metrics across the three protocols, grouped by the families introduced in Section~\ref{sec:setup}. Filtered rubric scoring achieves perfect rank agreement and the largest gap and spread, at a small drop in rubric-level unanimity.}
\label{tab:validation}
\renewcommand{\arraystretch}{1.3}
\setlength{\tabcolsep}{5pt}
\resizebox{\linewidth}{!}{%
\begin{tabular}{l c c c c c}
\toprule
\multirow{2}{*}{Protocol} & \multicolumn{2}{c}{\textbf{Rank stability}} & \textbf{Inter-judge agreement} & \multicolumn{2}{c}{\textbf{Discriminative power}} \\
\cmidrule(lr){2-3} \cmidrule(lr){4-4} \cmidrule(lr){5-6}
 & Spearman & Identical ranks & Unanimity (\%) & Avg gap & Spread \\
\midrule
Holistic scoring (WB-Score)    & 0.986 & \phantom{0}7/16 & --   & 2.24 & 33.5 \\
Rubric scoring, unfiltered     & 0.998 & 14/16           & 87.6 & 2.22 & 33.3 \\
Rubric scoring, filtered       & 1.000 & 16/16           & 83.8 & 3.26 & 48.9 \\
\bottomrule
\end{tabular}%
}
\end{table}

This rank invariance across judges has a direct practical consequence: a new candidate can be evaluated with any single judge. The cheapest in our suite, Gemini 3 Flash at approximately \$2.1 per candidate (Section~\ref{sec:setup}), reproduces the leaderboard obtained with the most expensive judge.

\subsection{Model ranking and analysis}\label{sec:results-ranking}

Table~\ref{tab:leaderboard} presents the final ranking on the effective benchmark using Sabi{\'a}-4 as judge; since the rank ordering is identical across the three judges (Section~\ref{sec:results-validation}), the substance of the leaderboard does not depend on this choice. Scores span from 88.6 (GPT-5.2) to 43.3 (Qwen2.5-7B), a 45.3-point spread.

\begin{table}[H]
\centering
\caption{Prosa leaderboard, Sabi{\'a}-4 as judge, filtered rubric scoring (0--100 scale).}
\label{tab:leaderboard}
\renewcommand{\arraystretch}{1.15}
\scalebox{1.0}{%
\begin{tabular}{@{}c l l r@{}}
\toprule
Rank & Model & Family & Score \\
\midrule
1  & GPT-5.2          & OpenAI   & 88.6 \\
2  & GPT-5 Mini       & OpenAI   & 84.8 \\
3  & Gemini 3 Pro     & Google   & 83.8 \\
4  & Qwen3-235B       & Alibaba  & 80.1 \\
5  & Gemini 3 Flash   & Google   & 78.8 \\
6  & Gemini 2.5 Flash & Google   & 76.3 \\
7  & GPT-4.1          & OpenAI   & 74.8 \\
8  & Qwen3-30B        & Alibaba  & 73.9 \\
9  & Sabi{\'a}-4          & Maritaca AI & 70.2 \\
10 & GPT-4.1 Mini     & OpenAI   & 67.7 \\
11 & Qwen3-4B         & Alibaba  & 64.6 \\
12 & Sabi{\'a}-3.1        & Maritaca AI & 60.5 \\
13 & GPT-4o           & OpenAI   & 55.0 \\
14 & GPT-4o Mini      & OpenAI   & 54.5 \\
15 & Qwen2.5-14B      & Alibaba  & 46.4 \\
16 & Qwen2.5-7B       & Alibaba  & 43.3 \\
\bottomrule
\end{tabular}%
}
\end{table}

Four observations stand out in Table~\ref{tab:leaderboard}. First, Qwen3-235B reaches rank 4 with a score of $80.1$, placing it above GPT-4.1 ($74.8$) and Gemini 3 Flash ($78.8$) and only $3.7$ points below Gemini 3 Pro, showing that an open-weights model is competitive with the proprietary frontier on natural Brazilian Portuguese chat. Second, the generational gap within the Qwen family is pronounced: the 2025-generation Qwen3 variants occupy ranks 4, 8, and 11, while the 2024-generation Qwen2.5 variants sit at ranks 15 and 16, with Qwen2.5-14B finishing below the 4B-parameter Qwen3-4B. Third, the GPT-4o generation is noticeably behind the GPT-4.1 and GPT-5 lines: GPT-4o ($55.0$) and GPT-4o Mini ($54.5$) land roughly $13$ points below GPT-4.1 Mini ($67.7$) and more than $33$ points below GPT-5.2. Finally, the Maritaca AI family shows a clear generational improvement on Brazilian Portuguese chat: Sabi{\'a}-4 reaches rank 9 with $70.2$, $9.6$ points above Sabi{\'a}-3.1 ($60.5$, rank 12).

\begin{figure}[!htbp]
    \centering
    \includegraphics[width=\linewidth]{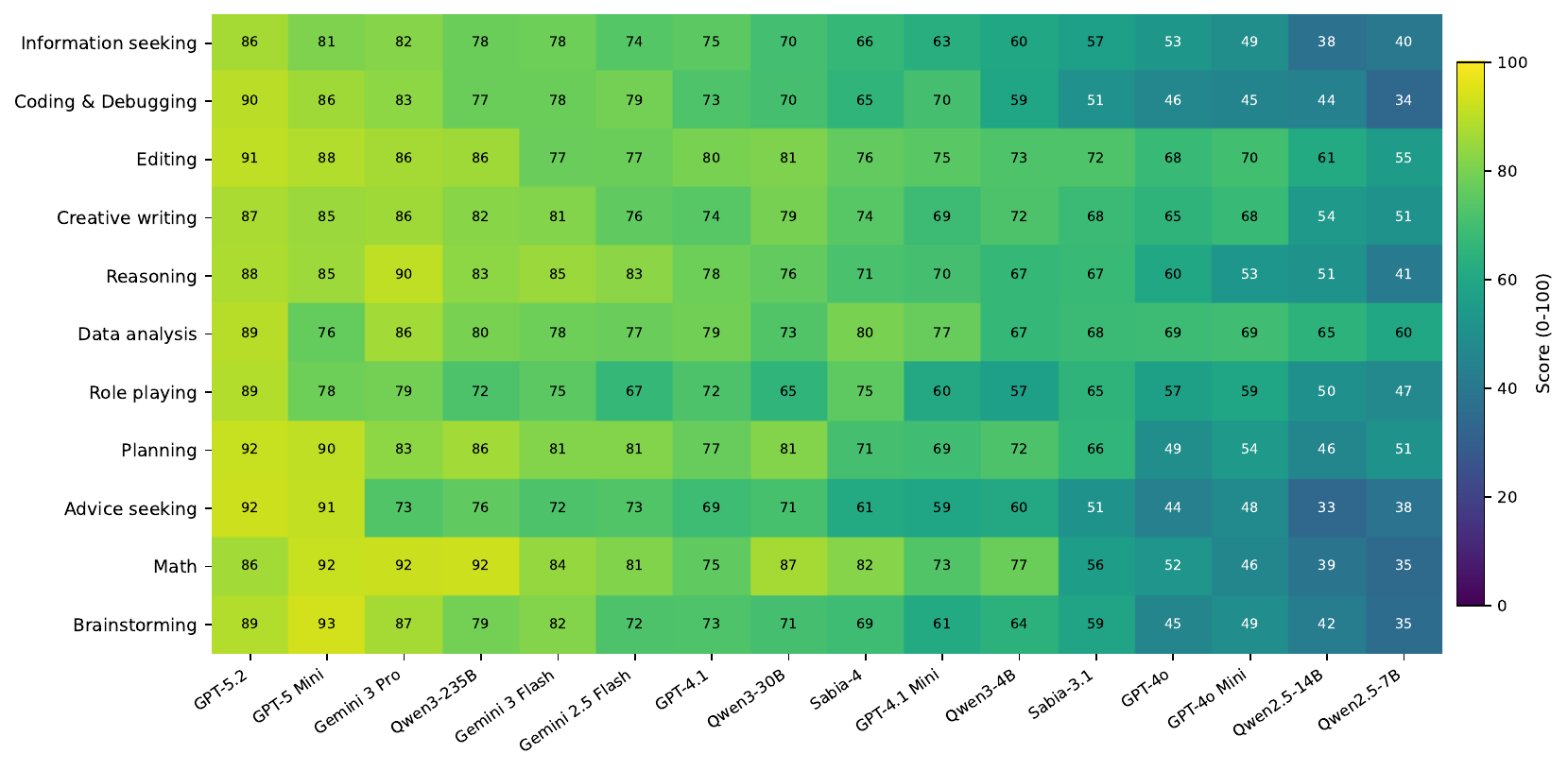}
    \caption{Per-category scores across the 11 topical categories of Prosa, with Sabi{\'a}-4 as judge under filtered rubric scoring (0--100 scale).}
    \label{fig:category-heatmap}
\end{figure}

Figure~\ref{fig:category-heatmap} breaks the ranking down by topical category. The benchmark discriminates most sharply on \emph{Advice seeking}, \emph{Brainstorming}, \emph{Math}, and \emph{Coding \& Debugging} (spreads of 55--60 points), and also on \emph{Reasoning} and \emph{Information seeking} (around 49 points). It discriminates least on \emph{Data analysis} (spread 29), where most current models produce similarly structured answers, and moderately on \emph{Editing} and \emph{Creative writing} (around 36 points each). The top candidates dominate every category, while the Qwen2.5 row shows a clear floor across the 11 topical categories.

\section{Conclusion}\label{sec:conclusion}

Prosa is the first real user multi-turn Brazilian Portuguese chat benchmark with rubric-based LLM-as-judge evaluation. Its central methodological lesson is that decomposing the judgement into atomic binary verdicts matters more than the choice of judge model: three judges from three model families agree on every one of the 16 candidate ranks under filtered binary rubric scoring, while the same judges disagree under holistic scoring. The multi-judge post-hoc filtering pipeline raises the average gap between neighbouring models by 47\%, sharpening Prosa's discriminative power.

Two consequences follow. First, rubric scoring lowers the cost barrier for benchmark evaluation: any reasonable judge produces the same ranking, the cheapest in our suite (Gemini 3 Flash at \$2.1 per candidate) reproduces the leaderboard of the most expensive, and benchmark designers can therefore swap in cheaper or future judges without re-baselining. Second, the per-rubric verdicts that yield judge-invariant rankings provide a more discriminative quality signal than a holistic score, which may make the same rubric set usable as a reward signal for reward modelling in addition to evaluation.

We release Prosa, the conversation filtering pipeline, the rubric-based scoring code, and the multi-judge rubric filtering pipeline. Because the judge's task of integrating multiple dimensions into a single scalar does not inherently depend on the prompt language, we conjecture that the central effect transfers beyond Brazilian Portuguese; validating this conjecture, running a human-alignment study, and applying the protocol to reward modelling are natural extensions.

\subsubsection{Ethics statement.}
Prosa is derived from the non-toxic release of WildChat~\cite{zhao2024wildchat} (WildChat-4.8M), distributed under the AI2 ImpACT Licenses. WildChat's authors removed personally identifiable information and hashed all IP addresses (mapped only to country and state) prior to release; we apply no further toxicity filter and inherit these protections. The conversations reflect real user interactions and may contain language that some readers find sensitive. Downstream use is subject to the AI2 ImpACT Licenses that govern WildChat.

\subsubsection{LLM usage disclosure.}
Language-model tools were used as writing aids for grammatical revision, clarity, and occasional style adjustments, and as coding aids throughout the codebase. They were not used for the formulation of research ideas, the literature review, the methodological choices, the analysis of empirical results, or the framing of the conclusions. All technical and scientific content was produced, reviewed, and validated by the authors, who assume full responsibility for the final text.

\subsubsection{\discintname}
The authors declare no conflicts of interest related to the content of this article.

\bibliographystyle{splncs04}
\bibliography{references}

\end{document}